\documentclass[runningheads]{llncs}

\usepackage{eccv}

\usepackage{eccvabbrv}

\usepackage{graphicx}
\usepackage{booktabs}
\usepackage{multirow}
\usepackage{multicol}
\usepackage[accsupp]{axessibility}  

\usepackage{eccvabbrv}
\usepackage[ruled,linesnumbered]{algorithm2e}

\usepackage{graphicx}
\usepackage{booktabs}

\usepackage[accsupp]{axessibility}  

\usepackage[pagebackref,breaklinks,colorlinks,citecolor=eccvblue]{hyperref}
\usepackage{multirow}
\usepackage{orcidlink}

\usepackage{xcolor}
\definecolor{mycolorpurple}{RGB}{255, 102, 255}
\definecolor{mycolorred}{RGB}{227, 26, 26}
\definecolor{mycolorskyblue}{RGB}{6, 185, 238}
\usepackage{threeparttable}

\begin{document}

\title{AHMF: Adaptive Hybrid-Memory-Fusion Model for Driver Attention Prediction} 
	
	\titlerunning{Abbreviated paper title}
	
	\author{
		Dongyang Xu\inst{1}\orcidlink{0009-0008-9927-8649} 
		\and
		Qingfan Wang\inst{1}\thanks{Corresponding author}\orcidlink{0000-0001-8556-1760} 
		\and
		Ji Ma\inst{2}\orcidlink{0009-0000-4110-113X} 
  		\and
		Xiangyun Zeng\inst{3}\orcidlink{} 
		\and
		Lei Chen\inst{3}\orcidlink{} 
		}

	\authorrunning{D.~Xu et al.}
	
	\institute{School of Vehicle and Mobility, Tsinghua University
		\and
		College of Engineering, Peking University
		\\
		\and
		SenseTime Research
		\\
	\email{\{xudy22\, wqf20\}@mails.tsinghua.edu.cn}
}
	\maketitle
\renewcommand{\thefootnote}{}
\footnotetext{D. Xu and Q. Wang --- Equal contributions.}
\footnotetext{D. Xu --- This work was done during the internship at SenseTime.}

\begin{abstract}
  Accurate driver attention prediction can serve as a critical reference for intelligent vehicles in understanding traffic scenes and making informed driving decisions. Though existing studies on driver attention prediction improved performance by incorporating advanced saliency detection techniques, they overlooked the opportunity to achieve human-inspired prediction by analyzing driving tasks from a cognitive science perspective. During driving, drivers' working memory and long-term memory play crucial roles in scene comprehension and experience retrieval, respectively. Together, they form situational awareness, facilitating drivers to quickly understand the current traffic situation and make optimal decisions based on past driving experiences. To explicitly integrate these two types of memory, this paper proposes an Adaptive Hybrid-Memory-Fusion (AHMF) driver attention prediction model to achieve more human-like predictions. Specifically, the model first encodes information about specific hazardous stimuli in the current scene to form working memories. Then, it adaptively retrieves similar situational experiences from the long-term memory for final prediction. Utilizing domain adaptation techniques, the model performs parallel training across multiple datasets, thereby enriching the accumulated driving experience within the long-term memory module. Compared to existing models, our model demonstrates significant improvements across various metrics on multiple public datasets, proving the effectiveness of integrating hybrid memories in driver attention prediction.
\end{abstract}

\section{Introduction}
\label{sec:intro}

Human drivers primarily rely on visual information to drive. The distribution of their visual attention reflects experienced drivers’ cognitive understanding of the current traffic scene, particularly in safety-critical scenarios with collision risks. For intelligent vehicles, accurately predicting driver attention is crucial for quickly identifying key risk elements in traffic scenes and assisting decision-making systems in making effective collision avoidance decisions \cite{hu2021utilising,karim2022dynamic}.

Due to such significant research significance, numerous studies on driver attention prediction have emerged \cite{deng2023driving,fu2023multimodal,lin2022ds,tian2022driving}. These studies typically adopted the basic encoder-decoder model architecture, using CNN or Transformer as core components. However, current attention prediction models’ performance improvements are primarily attributed to backbone advancements in computer vision while neglecting the necessary cognitive mechanism analysis of the driving task itself. Consequently, these models have not yet achieved human-inspired driver attention prediction.

During driving, human drivers must process complex and variable traffic information in real time, especially in safety-critical scenarios. This cognitive process involves both working memory and long-term memory \cite{wood2016working,broadbent2023cognitive}. The working memory module rapidly processes visual information by quickly identifying key risk objects in the current scene and assessing their danger \cite{zhang2023importance}. When a potential collision is imminent, drivers rapidly retrieve relevant experiences from long-term memory. Together, these processes help drivers form situational awareness, quickly comprehend the current traffic situation, and make optimal decisions based on accumulated driving experience \cite{krems2009driving,zhang2023importance,de2019situation,gan2021constructing}.

To achieve more human-like driver attention predictions, this paper proposes an Adaptive Hybrid-Memory-Fusion (AHMF) model by explicitly incorporating both working memory and long-term memory into driver attention prediction. Furthermore, leveraging domain adaptation, our model performs parallel training across multiple datasets, effectively enriching long-term memory with a diverse range of driving experiences. By combining specific dangerous stimuli in the scene (processed by the encoder as working memory) with retrieved experiences from the long-term memory, the model makes the final optimal predictions. We evaluated our model through comparative experiments on multiple public datasets. The results indicate that our model outperforms existing SOTA models across several metrics. The contributions of this paper are as follows:

\begin{enumerate}
\item We predict drivers’ visual attention in a manner that closely aligns with their understanding of traffic scenes from a cognitive science perspective. Specifically, the model first encodes specific dangerous stimuli in the current scene to form working memory, which is then integrated with long-term memory to produce the final scene encoding.

\item Utilizing domain adaptation, we achieve parallel training on multiple datasets, thereby enhancing the diversity of information in the long-term memory module and forming a comprehensive “driving experience” knowledge base, significantly boosting the model’s generalization capability.

\item Experiments demonstrate that our model achieves state-of-the-art prediction performance across several metrics on multiple public datasets.
\end{enumerate}
\section{Related Work}

\subsection{Driver Attention Prediction}

The research on driver attention prediction has evolved through three stages: early machine learning methods, CNN-based methods, and Transformer-based methods. Initially, classical machine learning approaches, such as dynamic Bayesian models, adopted bottom-up and top-down frameworks to simulate drivers’ visual attention \cite{pang2008stochastic,heracles2009dynamic,ban2010top}. With the development of CNN, convolutional prediction methods became mainstream. These models typically employ an encoder-decoder structure, where encoders process current scene information while decoders reconstruct visual attention distributions \cite{tawari2017computational,xia2019predicting,palazzi2018predicting,deng2019drivers,deng2021driving}. Recently, the impressive performance of Transformers in computer vision has led to the development of Transformer-based image/video saliency detection \cite{xie2022pyramid,huang2022temporally,ma2022video,chen2023fblnet}. Despite these advancements, these studies have not yet achieved human-inspired predictions, as they lack cognitive science insights into the driving task. Models that align more closely with human drivers’ scene understanding mechanisms are expected to enhance prediction accuracy further.

\subsection{Memory-Augmented Deep Learning}

The integrated development of cognitive science and deep learning has led to the birth of memory-augmented models that simulate external memory to overcome the limitations of working memory \cite{le2021memory}. A notable early example is the Long Short-Term Memory (LSTM) model \cite{hochreiter1997long}. Subsequent deep learning models have explored various forms of external memory integration \cite{lopez2017gradient,prakash2017condensed,gao2018motion,le2019learning,landi2021working}. For driver attention prediction, FBLNet incorporates a feedback loop structure to achieve incremental knowledge, which can be seen as a kind of simple long-term memory \cite{chen2023fblnet}. Our approach differs by proposing staged modeling of working memory and long-term memory and an effective memory fusion fashion. Meanwhile, incorporating domain adaptation significantly enriched the accumulated “driving experience” in long-term memory, as well as a better generalization ability.

\section{Method}

In this section, we proposed a novel Adaptive Hybrid-Memory-Fusion (AHMF) driver attention prediction model that explicitly incorporated drivers’ both working memory and long-term memory to achieve human-like predictions. Fig. 1 illustrates the overview of AHMF, which involves two core modules, \ie, temporal-spatial working memory encoding and attention-based hybrid memory fusion. In addition, necessary domain-specific modules were incorporated to enrich the accumulated long-term memory across various datasets. Given the length constraints of this paper, we will primarily use textual descriptions and avoid complex mathematical formulas to detail the proposed AHMF model.

\begin{figure}[tb]
  \centering
  \includegraphics[width=1\linewidth]{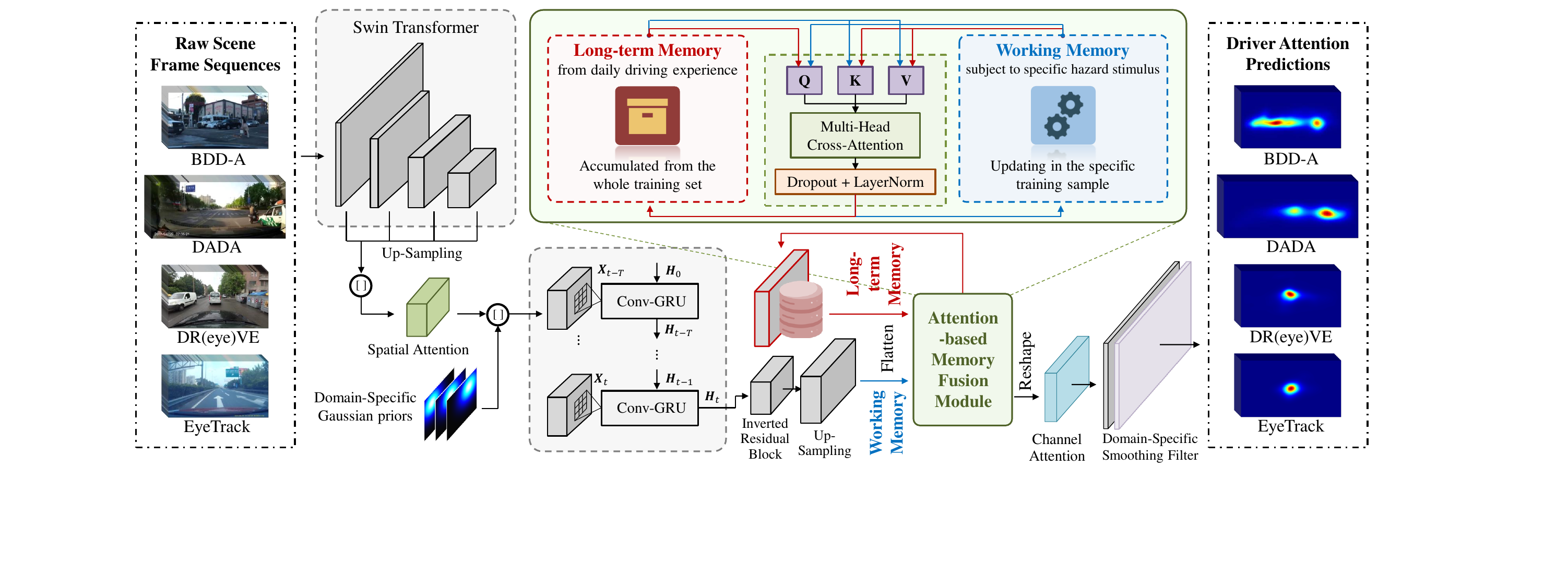}
  \caption{Overview of the proposed AHMF driver attention prediction model.}
  \label{fig:fig1}
\end{figure}

\subsection{Temporal-Spatial Working Memory Encoding}

Complex traffic scenes, particularly safety-critical scenarios, display not only strong temporal correlation characteristics but also spatially unbalanced distributions of critical risk objects, both significantly affecting the encoding process of drivers’ working memory. Thus, it is necessary to adopt the pattern of hybrid temporal-spatial encoding to achieve accurate driver attention prediction.

In terms of spatial encoding, we first adopted a Swin Transformer-tiny \cite{liu2021swin} as the backbone to extract essential semantic features. Extracted features at different levels were first normalized into the same dimension with up-sampling and then concatenated to capture semantic information in different ranges. Then, a spatial-attention-based convolution module was designed to model the inherent relationships over various local features of traffic frames, which has been proven effective in enhancing its spatial representation capability \cite{woo2018cbam,gan2021constructing}. The spatial attention module can be formulated as follows:
\begin{equation}
  \mathbf{y}_{i} = \sum_{j=1}^{H \times W} \left(W_{\omega} \mathbf{x}_{j}\right) \frac{\exp \left(\mathbf{x}_{i}^{T} W_{\theta}^{T} \cdot W_{\phi} \mathbf{x}_{j}\right)}{\sum_{j=1}^{H \times W} \exp \left(\mathbf{x}_{i}^{T} W_{\theta}^{T} \cdot W_{\phi} \mathbf{x}_{j}\right)}
  \label{eq:eq1}
\end{equation}
where $x_i$ and $y_i$ are the input and output pixel at the position index of $i$-th, respectively, $i, j \in \mathbb{R}^{(H \times W)}$ denote the flattened $1d$ position index of features along the spatial dimension, and the matrixes $W_{\theta}$, $W_{\phi}$, and $W_{\omega}$ represent the learnable parameters of $1 \mathrm{x}1$ convolution layers. Then, the encoded features were concatenated with predefined domain-specific Gaussian priors to enhance generalization ability in various heterogeneous driving scenes.

Regarding temporal encoding, we utilized convolutional gated recurrent units (Conv-GRUs) to realize the efficient transmission of temporal traffic information based on the update gate and reset gate \cite{shi2015convolutional,zhang2018detecting}. We selected Conv-GRUs instead of Conv-LSTM due to its lightweight network structure, which is more suitable for efficient online inference under safety-critical scenarios.

\subsection{Attention-based Hybrid Memory Fusion}
Human drivers’ working memory (for scene comprehension) and long-term memory (for experience retrieval) during driving were independently modeled in AHMF. Further, we designed an efficient adaptive memory fusion module using the attention mechanism, which was inspired by human drivers’ situational awareness mechanism.

\noindent
\textbf{Working Memory Modeling.}
Obtaining the temporal-spatial encoded features from the encoder, we used an inverted residual block \cite{sandler2018mobilenetv2} to reduce the channel dimension before adopting an up-sampling layer to adjust the spatial dimensions of features.

\noindent
\textbf{Long-term Memory Modeling.}
The long-term memory module was modeled as an offline knowledge base, and it was initialized as a set of learnable parameters with the same size as working memory (\ie, $H \times W \times C$). During training, it first retrieves critical driving experience according to the queries from working memory and is then updated to incorporate newly encoded features continuously.

\noindent
\textbf{Attention-based Hybrid Memory Fusion.}
We adopted two multi-head cross-attention-based fusion modules to facilitate the information transfer between two memories. Since attention modules take serialized data as inputs, working memory, and long-term memory were flattened at the spatial and channel dimensions.

First, to enhance the working memory with “driving experience” retrieved from long-term memories, we used linear layers to project working memory into queries and long-term memories into keys and values, respectively. A multi-head cross-attention module $\operatorname{MHCA(\cdot)}$ was adopted to model the inherent relationships between the two memories, developed as:
\begin{equation}
  Q_{w}=\operatorname{Linear}\left(m_{w}\right), \quad K_{l}=\operatorname{Linear}\left(m_{l}\right), \quad V_{l}=\operatorname{Linear}\left(m_{l}\right)
  \label{eq:eq2}
\end{equation}
\begin{equation}
m_{w}^{e}=\operatorname{MHCA}\left(q=Q_{w}+\operatorname{PE}\left(Q_{w}\right), k=K_{l}+\operatorname{PE}\left(K_{l}\right), v=V_{l}\right)
  \label{eq:eq3}
\end{equation}
where $m_w \in \mathbb{R}^{T \times (H \times W \times C)}$ and $m_l \in \mathbb{R}^{T \times (H \times W \times C)}$ represent the working memories and long-term memories, respectively, $m_{w}^{e} \in \mathbb{R}^{T \times H \times W \times C}$ denotes the enhanced working memories after reshaping to its original shape, $T$ is the sequence length, $H \times W \times C$ is the dimension of flattened features, $q$, $k$, and $v$ denote the query, key, and value in the multi-head cross-attention module, and $\operatorname{PE(\cdot)}$ denotes the sinusoidal position encoding of input tokens. Dropout and layer normalization were used after the cross-attention to stabilize the training process and avoid over-fitting.

Another multi-head cross-attention module with the switched keys, values (\ie, $K_w$ and $V_w$ projected from working memory), and queries (\ie, $Q_l$ projected from long-term memories) was used to update the accumulated long-term memories (from $m_{l}$ to $m_{l}^{e}$) with the newly encoded working memory $m_{w}^{e}$. Long-term memories benefited from various driving experiences across multiple datasets. When used for online inference, the cross-attention module for updating long-term memories can be deprecated to accelerate inference time.

After memory fusion, a channel-attention-based convolution module enhanced the representation of high-level features across channels, formulated as: 
\begin{equation}
  \boldsymbol{y}_{i}=\sum_{j=1}^{H \times W} \frac{\exp \left(\boldsymbol{x}_{i}^{T} \cdot \boldsymbol{x}_{\boldsymbol{j}}\right)}{\sum_{j=1}^{H \times W} \exp \left(\boldsymbol{x}_{\boldsymbol{i}}^{T} \cdot \boldsymbol{x}_{j}\right)}\left(\boldsymbol{x}_{j}\right)
  \label{eq:eq4}
\end{equation}
where dot-product similarity $x_{i}^{T}\cdot x_{j}$ measures the mutual influences between $i$-th channel and $j$-th channel of input features. Finally, smoothing filters were used to blur final attention predictions.


\subsection{Domain-Specific Modules for Domain Adaption}
For better generalization ability, we also incorporated a series of domain adaptation techniques to perform parallel training across multiple datasets, which can further enrich the accumulated driving experience within the long-term memories. Following the previous studies on video saliency modeling \cite{droste2020unified,gan2022multisource}, three domain-specific modules were utilized in AHMF, including domain-specific batch normalization, domain-specific Gaussian priors, and domain-specific smoothing filters. \textbf{Domain-specific batch normalization} aims at reducing undesirable data heterogeneity across different datasets during batch normalization \cite{chang2019domain}. \textbf{Domain-specific Gaussian priors}, modeled by a series of learnable parameters, serve as vital spatial prior information for driver attention prediction for a specific dataset \cite{deng2016does}. Considering the heterogeneous sharpness distributions of attention maps across datasets, we adopted \textbf{ domain-specific smoothing filters} to blur final attention predictions to improve performance \cite{gan2022multisource}.

\section{Experiments}
In this section, we describe our experimental setup and compare the proposed AHMF model with several SOTA methods. For implementation details of AHMF, please refer to the supplementary material (S.1.1).

\subsection{Experimental Settings}
\textbf{Datasets.}
The AHMF model was jointly trained and tested on four widely-used public large-scale driver attention datasets, \ie, Driver Attention and Driver Accident (DADA) \cite{fang2019dada}, Berkeley DeepDrive attention (BDD-A) \cite{xia2019predicting}, DReyeVE \cite{alletto2016dr}, and EyeTrack \cite{deng2019drivers}. Detailed descriptions of these datasets can be seen in the supplementary material (S.1.2).

\noindent
\textbf{Evaluation Metrics.}
To comprehensively evaluate model performance, we adopt various saliency evaluation metrics, including three distribution-based metrics, \ie, Similarity (SIM), Kullback-Leibler divergence (KLD), and Pearson’s correlation coefficient (CC), and two location-based metrics, \ie, Normalized Scanpath Saliency (NSS) and Area Under ROC Curve Judd (AUC-J). Details are given in the supplementary material (S.1.3).

\begin{footnotesize}
\begin{table}[t]
    \begin{center}
        \caption{Comparison of the AHMF model with state-of-the-art driver attention prediction models on the DADA and BDD-A datasets. Red values indicate the best performance, while blue values indicate the second-best.}
         \label{tab:tab1}
         \setlength{\tabcolsep}{3pt}
          \resizebox{0.95\linewidth}{!}{
              \begin{tabular}{ccccccccccc}
                \toprule
                \multirow{2}{*}{\raisebox{-0.7mm}[0pt][0pt]{Method}} & \multicolumn{5}{c}{DADA\cite{fang2019dada}} & \multicolumn{5}{c}{BDD-A\cite{xia2019predicting}} \\
                 \cmidrule(r){2-6}
                 \cmidrule(r){7-11}
                 &  AUC-J$\uparrow$ & SIM$\uparrow$ & CC$\uparrow$ & KLD$\downarrow$  & NSS$\uparrow$ &  AUC-J$\uparrow$ & SIM$\uparrow$ & CC$\uparrow$ & KLD$\downarrow$  & NSS$\uparrow$   \\ 
                \midrule
                {BDDA\cite{xia2019predicting}}
                      & 0.89 & 0.22 & 0.33 & 2.77 & 2.52 & \textbf{\textcolor{blue}{0.93}} & 0.35 & 0.48 & 2.07 & 3.45\\
                {U2NET\cite{qin2020u2}}
                      & \textbf{\textcolor{blue}{0.94}} & 0.30 & \textbf{\textcolor{blue}{0.47}} & 1.85 & 3.77 & \textbf{\textcolor{red}{0.95}} & 0.36 & 0.55 & 1.47 & 3.95 \\ 
                {MINET\cite{pang2020multi}}
                      & 0.86 & 0.30 & 0.38 & 9.99 & 3.62 & 0.86 & 0.35 & 0.48 & 10.5 & 4.28 \\ 
                {DRIVER\cite{bao2021drive}}
                      & 0.90 & 0.24 & 0.37 & 4.03 & 3.06 & 0.76 & 0.25 & 0.31 & 13.8 & 2.56 \\ 
                {ADA} \cite{gan2022multisource}
                      & \textbf{\textcolor{blue}{0.94}} & 0.36 & \textbf{\textcolor{red}{0.50}} & \textbf{\textcolor{blue}{1.59}} & 3.51 & 0.92 & \textbf{\textcolor{blue}{0.49}} & \textbf{\textcolor{red}{0.64}} & \textbf{\textcolor{red}{1.02}} & 4.56 \\
                {DADA\cite{fang2019dada}}
                      & \textbf{\textcolor{red}{0.95}} & 0.34 & \textbf{\textcolor{blue}{0.47}} & 2.16 & 4.05 & \textbf{\textcolor{red}{0.95}} & 0.39 & 0.55 & 1.48 & 4.22 \\
                {DBNET\cite{tian2022driving}}
                      & 0.91 & 0.25 & 0.39 & 2.77 & 2.93 & \textbf{\textcolor{red}{0.95}} & 0.39 & 0.55 & 1.85 & 3.93 \\  
                {PGNet\cite{xie2022pyramid}}
                      & 0.92 & \textbf{\textcolor{blue}{0.37}} & 0.45 & 5.28 & 4.04 & 0.92 & 0.43 & 0.56 & 6.09 & 4.92 \\
                {FBLNet\cite{chen2023fblnet}}
                      & \textbf{\textcolor{red}{0.95}} & 0.33 & \textbf{\textcolor{red}{0.50}} & 1.92 & \textbf{\textcolor{blue}{4.13}} & \textbf{\textcolor{red}{0.95}} & 0.46 & \textbf{\textcolor{red}{0.64}} & 1.40 & \textbf{\textcolor{red}{5.02}} \\
                \cmidrule(r){1-11}
                {\textbf{Ours}}
                      & \textbf{\textcolor{blue}{0.94}} & \textbf{\textcolor{red}{0.57}} & \textbf{\textcolor{red}{0.50}} & \textbf{\textcolor{red}{1.55}} & \textbf{\textcolor{red}{5.02}} & 0.92 & \textbf{\textcolor{red}{0.75}} & \textbf{\textcolor{blue}{0.58}} & \textbf{\textcolor{blue}{1.17}} & \textbf{\textcolor{red}{5.51}} \\
                \bottomrule
              \end{tabular}
          }
    \end{center}
\end{table}
\end{footnotesize}

\subsection{Comparison with SOTA}

\textbf{Quantitative Results.}
Compared with several SOTA driver attention prediction models, our model demonstrates satisfying performance on both the DADA and BDD-A datasets (Table \ref{tab:tab1}). Specifically, our model performs particularly well in SIM and NSS, achieving 0.57 and 5.02 on DADA and 0.75 and 5.51 on BDD-A, respectively, yielding a significant improvement of $+$54.0\%, $+$21.5\%, $+$53.1\%, and $+$9.8\% compared to the best existing methods. Additionally, our model achieves the lowest KLD of 1.55 and the highest CC of 0.50 on the DADA dataset. Furthermore, our model performs competitively in AUC-J and CC, closely matching the top-performing models. Overall, the AHMF model demonstrates significant advantages across multiple metrics, showing its robustness and effectiveness in driver attention modeling.



\noindent
\textbf{Qualitative Results.} Figure \ref{fig:fig2} visualizes the comparison results of driver attention prediction. The MLNet model overly focuses on static areas in video frames, unlike human drivers who concentrate on objects with specific inter-frame changes. Similarly, PGNet not only pays too much attention to static regions but also fails to allocate attention to the road ahead in the absence of salient objects, dispersing attention to irrelevant areas such as roadside fences. In contrast, our model focuses on dynamic traffic participants with significant inter-frame changes, and its attention predictions align closely with the ground truth. More qualitative results can be found in the supplementary material (S.2.1).

\begin{figure}[tb]
  \centering
  \includegraphics[width=1\linewidth]{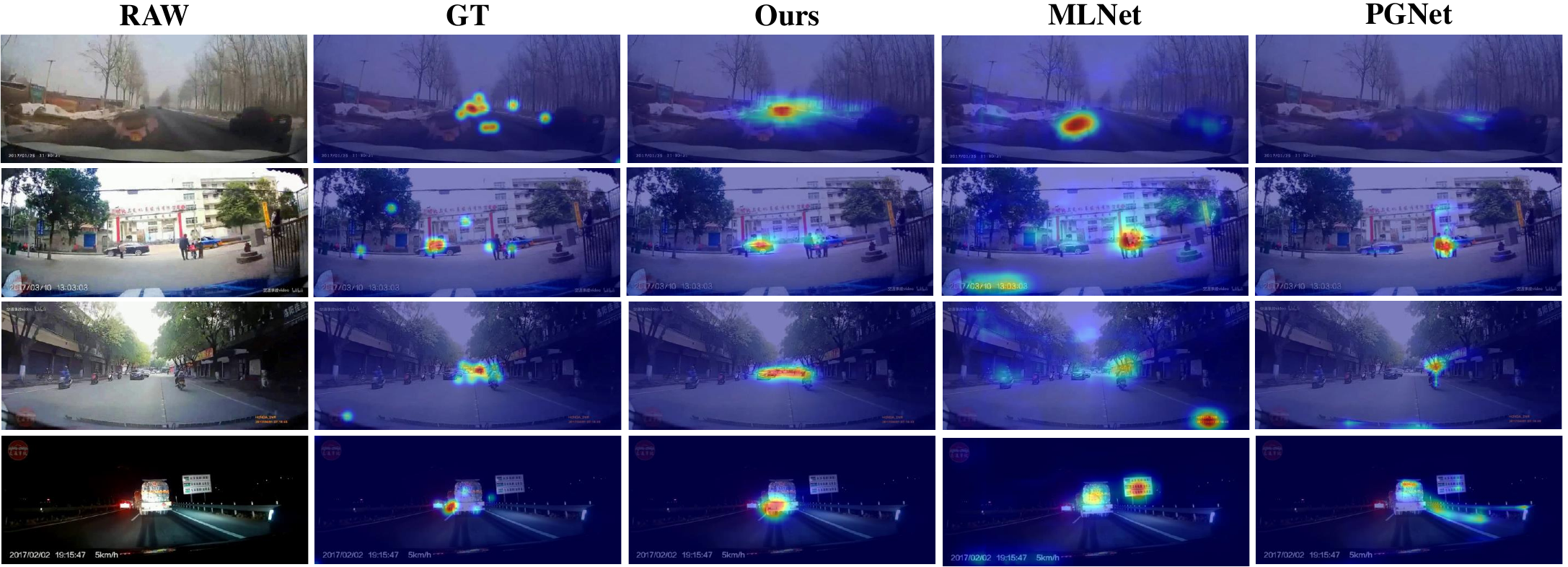}
  \caption{Qualitative results of the predicted driver attention maps. From left to right: raw inputs, ground-truth maps, predictions of ours, MLNet \cite{cornia2016deep}, and PGNet \cite{wang2021pgnet}.}
  \label{fig:fig2}
\end{figure}

\begin{table}[t]
  \centering
  \begin{minipage}[t]{0.48\textwidth}
    \centering
    \caption{Ablation study for AHMF's different components on the DADA dataset.}
    \label{tab:tab3}
        \resizebox{1.0\linewidth}{!}{
            \begin{tabular}{cccccc}
            \toprule
             & AUC-J$\uparrow$ & SIM$\uparrow$ & CC$\uparrow$ & KLD$\downarrow$  & NSS$\uparrow$ \\
            \midrule
            {w/o HMF}
                    & 0.92 & 0.44 & 0.47 & 1.61 & 4.49\\  
            {w/o SA}
                   & 0.92 & 0.56 & 0.49 & 1.56 & 4.95\\ 
            {w/o CA}
                   & 0.93 & 0.53 & 0.48 & 1.56 & 4.83\\ 
            \cmidrule(r){1-6}
            {\textbf{w/ all}}
                   & \textbf{0.94} & \textbf{0.57} & \textbf{0.50} & \textbf{1.55} & \textbf{5.02}\\ 
            \bottomrule
          \end{tabular}
        }
\end{minipage}\hfill
  \begin{minipage}[t]{0.48\textwidth}
    \centering
    \caption{Ablation study for the duration of working memory on DADA.}
    \label{tab:tab4}
        \resizebox{0.89\linewidth}{!}{
            \begin{tabular}{cccccc}
            \toprule
             & AUC-J$\uparrow$ & SIM$\uparrow$ & CC$\uparrow$ & KLD$\downarrow$  & NSS$\uparrow$ \\
            \midrule
            {1.0 s}
                   & 0.92 & 0.53 & 0.48 & 1.57 & 4.79\\ 
            {1.5 s}
                   & 0.92 & 0.55 & 0.48 & 1.55 & 4.86\\ 
                   
            {2.0 s}
                   & 0.92 & 0.57 & 0.47 & 1.58 & 4.81\\ 
    
           \cmidrule(r){1-6} 
            {\textbf{3.0 s}}
                   & \textbf{0.94} & \textbf{0.57} & \textbf{0.50} & \textbf{1.55} & \textbf{5.02}\\ 
            \bottomrule
        \end{tabular}
    }
  \end{minipage}
\end{table}

\subsection{Ablation Study}
We now analyze several design choices for AHMF in a series of ablation studies on the DADA dataset.

Table \ref{tab:tab3} presents the results of an ablation study evaluating the contributions of different key components of our model. Specifically, it compares the performance of the whole model against three variants: without the  Hybrid-Memory-Fusion (HMF) module, without the Spatial-Attention (SA) module, and without the Channel-Attention (CA) module. The results indicate that removing any component results in decreased performance, demonstrating the critical importance of each component in achieving optimal performance. As expected, the introduction of the HMF module shows the most significant improvement, increasing SIM by $+$29.5\% and NSS by $+$11.8\%.

Table \ref{tab:tab4} analyzes the impacts of working memory durations in traffic scenes, \ie, the length of historical input sequences. The results indicate that the shorter memory durations result in lower performance, particularly noticeable in the NSS metric. This demonstrates longer input sequences contribute to the temporal encoding of working memories and achieve more effective fusion of hybrid memories, finally enhancing driver attention predictions. 

More detailed ablation experiments can be found in the supplementary material (S.2.2).

\section{Conclusion}
In this paper, we have presented a novel, more human-like driver attention prediction model that incorporates both working memory and long-term memory. Unlike the existing approaches, our method explicitly modeled human drivers’ working memory for scene comprehension and long-term memory for experience retrieval to imitate their situational awareness mechanism when locating their visual attention during driving. Experiments proved that the proposed method of memory modeling and fusion significantly contributed to the performance improvement of driver attention prediction. We modeled the two kinds of memories in a very straightforward way. More future efforts should be made to find a better method of memory modeling with a more in-depth interdisciplinary study between cognitive science and computer vision.

\clearpage

\bibliographystyle{unsrt}
\bibliography{main}

\clearpage

\appendix
\section{Experimental Settings}

\subsection{Implementation Details.}
For the four datasets, we followed the data processing rules in \cite{gan2022multisource}. The video split rule followed the original datasets for a fair comparison. {The initialized learning rate of $0.01$ decayed as the exponential equation by a constant of $0.8$ after every epoch. Stochastic Gradient Descent with the momentum of $0.9$ and weight decay of $10^{-4}$ was employed to optimize the network. We froze the weight of the Swin transformer trained on the ImageNet dataset.  Batch size was set to $4$ for all datasets. We set the joint training iterations to $16$ epochs for all datasets. The model training was stopped as soon as the performance on the DADA validation set consecutively decreased three times than the performance at the prior training epoch. Our model was implemented using the PyTorch framework and trained on two NVIDIA A800 GPUs.}

\subsection{Datasets.}
Our proposed AHMF model was jointly trained on four available driver attention datasets, \ie, BDD-A \cite{xia2019predicting}, DADA \cite{fang2019dada}, DReyeVE \cite{alletto2016dr}, and EyeTrack \cite{deng2019drivers}. The Berkeley DeepDrive attention (BDD-A) dataset was filtered from the BDDV dataset, focusing on braking events in heavy traffic. Forty-five gaze providers labeled 1,429 critical scene videos. The Driver Attention and Driver Accident (DADA) dataset includes 2,000 accident videos across 54 kinds of accident scenarios collected from public websites worldwide. Twenty volunteers annotated visual attention points using the infrared eye tracker under lab conditions. DReyeVE was the first large-scale public driver attention dataset, comprising 74 traffic video clips collected from naturalistic driving experiments. EyeTrack’s traffic videos were collected by a dashcam on urban highways in China, and gaze data were captured in-lab as 28 subjects viewed the recorded clips.

\subsection{Evaluation Metrics.}
To comprehensively evaluate the performance of our proposed method, we adopt various saliency evaluation metrics, including three distribution-based metrics, \ie, Similarity (SIM), Kullback-Leibler divergence (KLD), and Pearson’s correlation coefficient (CC), and two location-based metrics, \ie, Normalized Scanpath Saliency (NSS) and Area Under ROC Curve Judd (AUC-J), calculated as:
\begin{footnotesize}  
\begin{equation}
  K L D(S, \hat{S})=\sum_{i=1}^{N} S(i) \log \left(\varepsilon+\frac{S(i)}{\varepsilon+\hat{S}(i)}\right)
  \label{eq:eq1}
\end{equation}
\end{footnotesize}
\begin{footnotesize}  
\begin{equation}
   \quad C C(S, \hat{S})=\frac{\operatorname{cov}(S, \hat{S})}{\sigma(S) \cdot \sigma(\hat{S})}
  \label{eq:eq2}
\end{equation}
\end{footnotesize}
\begin{footnotesize} 
\begin{equation}
  \operatorname{SIM}(S, \hat{S})=\sum_{i=1}^{N} \min (S(i), \hat{S}(i))
  \label{eq:eq3}
\end{equation}
\end{footnotesize}
\begin{footnotesize} 
\begin{equation}
   \quad N S S(P, \hat{S})=\frac{1}{\sum_{i=1}^{N} P_{i}} \times \sum_{i=1}^{N} P_{i} \frac{\hat{S}-\mu_{\hat{S}}}{\sigma_{\hat{S}}}
  \label{eq:eq4}
\end{equation}
\end{footnotesize}
where $S$ and $\hat{S}$ are the ground-truth and predicted saliency maps, respectively, $P$ is the ground-truth fixation point map, index $i$ denotes the $i-th$ pixel across all $N$ pixels.

\begin{figure}[h]
  \centering
  \includegraphics[width=1\linewidth]{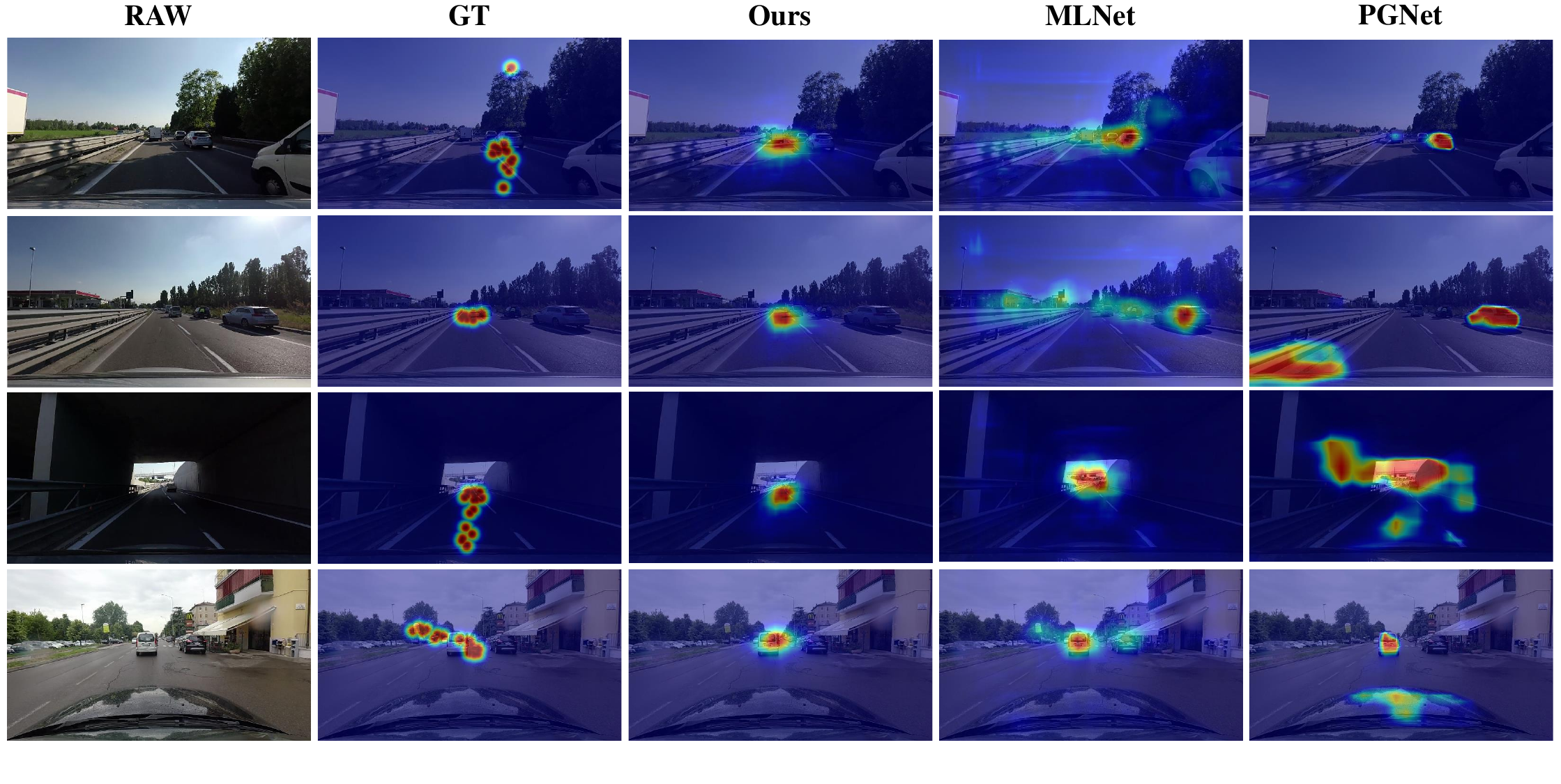}
  \caption{Qualitative results of the predicted driver attention maps in the DReyeVE dataset. From left to right: raw inputs, ground-truth attention maps, predictions of ours, MLNet\cite{cornia2016deep}, and PGNet \cite{wang2021pgnet}. }
  \label{fig:fig1}
\end{figure}
\begin{figure}[h]
  \centering
  \includegraphics[width=1\linewidth]{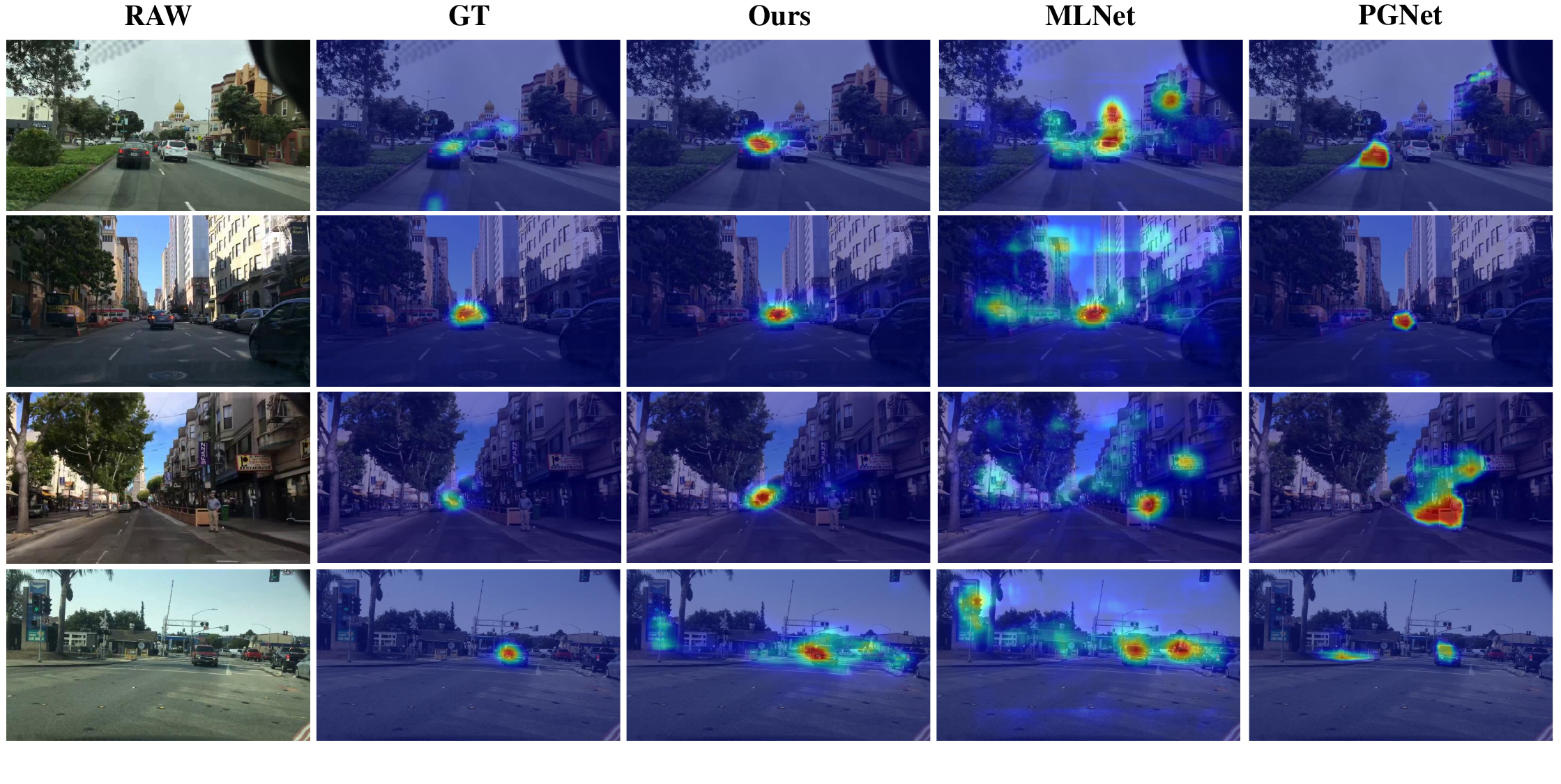}
  \caption{Qualitative results of the predicted driver attention maps in the BDD-A dataset. From left to right: raw inputs, ground-truth attention maps, predictions of ours, MLNet\cite{cornia2016deep}, and PGNet \cite{wang2021pgnet}. }
  \label{fig:fig2}
\end{figure}

\section{Results}
\subsection{Qualitative Results}
As illustrated in Figure \ref{fig:fig1}, the qualitative results demonstrate the superiority of our method when tested on the DReyeVE dataset.   In the video frame sequence tests on highways, the MLNet model tends to disperse part of its attention to stationary vehicles in adjacent lanes, whereas human drivers primarily focus their attention on the road ahead.  Similarly, the PGNet model not only disperses attention but also fails to handle vehicle reflections adequately, leading to an excessive focus on the host vehicle.   Additionally, PGNet performs poorly in scenarios with contrasting lighting conditions, such as tunnels.   In contrast, our model more accurately simulates human attention distribution, with its attention prediction results being highly consistent with the ground truth.

The results in Figure \ref{fig:fig2} illustrate the comparative performance of our proposed method against several baseline approaches on the BDD-A dataset.  Notably, MLNet's attention distribution frequently targets vehicles or objects situated outside the driving lane.  In scenarios devoid of prominent vehicles, both MLNet and PGNet exhibit a propensity to divert their attention towards roadside objects.  Conversely, our model more accurately replicates human attention distribution, with its predicted attention maps exhibiting a high degree of congruence with the ground truth.  These findings underscore the superior capability of our model to emulate driver attention, demonstrating enhanced focus on regions pertinent to the driving task.

\newpage

\begin{footnotesize}
\begin{table}[t]
    \begin{center}
        \caption{Ablation study for different components of AHMF on DADA and BDD-A dataset.}
         \label{tab:tab1}
         \setlength{\tabcolsep}{3pt}
          \resizebox{1.0\linewidth}{!}{
              \begin{tabular}{ccccccccccc}
                \toprule
                \multirow{2}{*}{\raisebox{-0.7mm}[0pt][0pt]{Method}} & \multicolumn{5}{c}{DADA\cite{fang2019dada}} & \multicolumn{5}{c}{BDD-A\cite{xia2019predicting}} \\
                 \cmidrule(r){2-6}
                 \cmidrule(r){7-11}
                 &  AUC-J$\uparrow$ & SIM$\uparrow$ & CC$\uparrow$ & KLD$\downarrow$  & NSS$\uparrow$ &  AUC-J$\uparrow$ & SIM$\uparrow$ & CC$\uparrow$ & KLD$\downarrow$  & NSS$\uparrow$   \\ 
                \midrule
                    \textbf{AHMF} & \textbf{0.94} & \textbf{0.57} & \textbf{0.50} & \textbf{1.55} & \textbf{5.02} & \textbf{0.92} & \textbf{0.75} & \textbf{0.58} & \textbf{1.17} & \textbf{5.51} \\
                    -SA &       0.92 & 0.56 & 0.49 & 1.56 & 4.95 & 0.91 & 0.75 & 0.57 & 1.21 & 5.43 \\
                    -CA &       0.93 & 0.53 & 0.48 & 1.56 & 4.83 & 0.91 & 0.75 & 0.56 & 1.20 & 5.44 \\
                    -HMF &       0.92 & 0.44 & 0.47 & 1.61 & 4.49 & 0.91 & 0.61 & 0.57 & 1.19 & 4.94 \\
                \bottomrule
              \end{tabular}
          }
    \end{center}
\end{table}
\end{footnotesize}

\begin{table}[ht]
    \begin{center}
        \caption{Ablation study for the position of the accumulation long-term memories.}
         \label{tab:tab2}
         \setlength{\tabcolsep}{3pt}
          \resizebox{1.0\linewidth}{!}{
              \begin{tabular}{ccccccccccc}
                \toprule
                \multirow{2}{*}{\raisebox{-0.7mm}[0pt][0pt]{Location}} & \multicolumn{5}{c}{DADA\cite{fang2019dada}} & \multicolumn{5}{c}{BDD-A\cite{xia2019predicting}} \\
                 \cmidrule(r){2-6}
                 \cmidrule(r){7-11}
                 &  AUC-J$\uparrow$ & SIM$\uparrow$ & CC$\uparrow$ & KLD$\downarrow$  & NSS$\uparrow$ &  AUC-J$\uparrow$ & SIM$\uparrow$ & CC$\uparrow$ & KLD$\downarrow$  & NSS$\uparrow$   \\ 
                \midrule
                    After HMF & \textbf{0.94} & \textbf{0.57} & \textbf{0.50} & \textbf{1.55} & \textbf{5.02} & \textbf{0.92} & \textbf{0.75} & \textbf{0.58} & 1.17 & \textbf{5.51} \\
                    After CA & 0.92 & 0.54 & 0.49 & \textbf{1.55} & 4.98 & 0.91 & 0.72 & \textbf{0.58} & \textbf{1.16} & 5.36 \\
                    
                \bottomrule
              \end{tabular}
          }
    \end{center}
\end{table}

\begin{table}[ht]
    \begin{center}
        \caption{Ablation study for backbones. SWT represents Swin-Transformer-Tiny. }
         \label{tab:tab3}
            \resizebox{0.95\linewidth}{!}{
                \begin{tabular}{ccccccccccccc}
                \toprule
                \multirow{2}{*}{\raisebox{-0.7mm}[0pt][0pt]{Method}} & 
                \multirow{2}{*}{\raisebox{-0.7mm}[0pt][0pt]{Paras}} &
                \multirow{2}{*}{\raisebox{-0.7mm}[0pt][0pt]{FLOPS}} &
                \multicolumn{5}{c}{DADA\cite{fang2019dada}} & \multicolumn{5}{c}{BDD-A\cite{xia2019predicting}} \\
                 \cmidrule(r){4-8}
                 \cmidrule(r){9-13}
                 &&&  AUC-J$\uparrow$ & SIM$\uparrow$ & CC$\uparrow$ & KLD$\downarrow$  & NSS$\uparrow$ &  AUC-J$\uparrow$ & SIM$\uparrow$ & CC$\uparrow$ & KLD$\downarrow$  & NSS$\uparrow$   \\ 
                \midrule
                {SWT}
                       & 28.2M & 4.38G & \textbf{0.94} & \textbf{0.57} & \textbf{0.50} & \textbf{1.55} & \textbf{5.02} & \textbf{0.92} & \textbf{0.75} & \textbf{0.58} & \textbf{1.17} & \textbf{5.51}\\  
                {Resnet-50}
                       & 25.6M & 4.13G & 0.92 & 0.52 & 0.47 & 1.60 & 4.76 & 0.91 & 0.73 & \textbf{0.58} & 1.18 & 5.43\\
                \bottomrule
              \end{tabular}
            }
    \end{center}
\end{table}

\begin{footnotesize}
\begin{table}[t]
    \begin{center}
        \caption{Ablation study on the length of historical input sequences.}
         \label{tab:tab4}
         \setlength{\tabcolsep}{3pt}
          \resizebox{1.0\linewidth}{!}{
              \begin{tabular}{ccccccccccc}
                \toprule
                \multirow{2}{*}{\raisebox{-0.7mm}[0pt][0pt]{Length}} & \multicolumn{5}{c}{DADA\cite{fang2019dada}} & \multicolumn{5}{c}{BDD-A\cite{xia2019predicting}} \\
                 \cmidrule(r){2-6}
                 \cmidrule(r){7-11}
                 &  AUC-J$\uparrow$ & SIM$\uparrow$ & CC$\uparrow$ & KLD$\downarrow$  & NSS$\uparrow$ &  AUC-J$\uparrow$ & SIM$\uparrow$ & CC$\uparrow$ & KLD$\downarrow$  & NSS$\uparrow$   \\ 
                \midrule
                    30 & \textbf{0.94} & \textbf{0.57} & \textbf{0.50} & \textbf{1.55} & \textbf{5.02} & \textbf{0.92} & 0.75 & \textbf{0.58} & 1.17 & \textbf{5.51} \\
                    20 & 0.92 & \textbf{0.57} & 0.47 & 1.58 & 4.81 & 0.91 & 0.75 & \textbf{0.58} & 1.14 & 5.42 \\
                    15 & 0.92 & 0.55 & 0.48 & \textbf{1.55} & 4.86 & 0.91 & 0.73 & 0.57 & \textbf{1.12} & 5.47 \\
                    10 & 0.92 & 0.53 & 0.48 & 1.57 & 4.79 & 0.91 & 0.72 & \textbf{0.58} & 1.14 & 5.37 \\
                    5  & 0.92 & 0.56 & 0.47 & 1.60 & 4.65 & 0.91 & \textbf{0.79} & 0.57 & 1.23 & 5.35 \\   
                \bottomrule
              \end{tabular}
          }
    \end{center}
\end{table}
\end{footnotesize}
\subsection{Ablation Study}
Table \ref{tab:tab1} presents the results of an ablation study evaluating the contributions of different components of AHMF on the DADA and BDD-A datasets. The full AHMF model is compared against three variants: without the Self-Attention (SA) module, without the Channel-Attention (CA) module, and without the Hybrid-Memory-Fusion (HMF) module. The results on both datasets indicate that each component contributes to the model's effectiveness, with the combination of all components yielding the optimal performance. Among the three core components, removing the HMF module causes the most significant performance degradation, reducing NSS by -10.6\% and -10.3\% on the DADA and BDD-A datasets, respectively. This demonstrates the importance of introducing the long-term memory module and designing an appropriate hybrid memory fusion module for driver attention prediction tasks.


Table \ref{tab:tab2} displays an ablation study for the position of the long-term memory updating module. The results show that updating long-term memories after the HMF module results in slightly better performance across most metrics compared to updating it after CA. This discrepancy can be attributed to the nature of the feature information processing in these stages. After the HMF phase, the amalgamation of feature information enables the model to more effectively leverage memory, thereby optimizing performance.  Conversely, after the CA module, the feature information tends to become more abstract and hierarchical, potentially disrupting the initial memory sequence and leading to a drop in performance.

Table \ref{tab:tab3} demonstrates that using Swin-Transformer-Tiny (SWT) as the backbone outperforms Resnet-50 on both the DADA and BDD-A datasets.

The ablation study in Table \ref{tab:tab4} shows that using longer historical input sequences significantly enhances model performance across various metrics for both the DADA and BDD-A datasets. The highest AUC-J, SIM, and NSS scores are achieved with a length of 30, indicating that capturing more temporal information allows the model to better encode working memories to fully understand current traffic scenarios and potentially conflicting objects. Additionally, longer sequences also reduce noise and contribute to more robust predictions.





\clearpage  

%
%

\end{document}